\documentclass[10pt,twocolumn,letterpaper]{article}

\usepackage{iccv}
\usepackage{times}
\usepackage{epsfig}
\usepackage{graphicx}
\usepackage{amsmath}
\usepackage{amssymb}

\usepackage[pagebackref=true,breaklinks=true,letterpaper=true,colorlinks,bookmarks=false]{hyperref} 

\usepackage{subcaption} 
\usepackage{framed,multirow}

\usepackage{latexsym}
\usepackage{pifont}
\usepackage{caption}
\DeclareMathOperator*{\argmin}{arg\,min}

\newtheorem{theorem}{Theorem}
\newtheorem{lemma}{Lemma}
\newtheorem{definition}{Definition} 

\iccvfinalcopy 

\ificcvfinal\fi

\begin{document}

\title{GeomNet: A Neural Network Based on Riemannian Geometries of SPD Matrix Space and Cholesky Space for 3D Skeleton-Based Interaction Recognition}  

\author{Xuan Son Nguyen\\
ETIS UMR 8051, CY Cergy Paris Universit\'e, ENSEA, CNRS, F-95000, Cergy, France\\
{\tt\small xuan-son.nguyen@ensea.fr}
}

\maketitle
\ificcvfinal\thispagestyle{empty}\fi

\begin{abstract}
In this paper, we propose a novel method for representation and classification of two-person interactions from 3D skeleton sequences. 
The key idea of our approach is to use Gaussian distributions to capture statistics on $\mathbb{R}^n$ and those on the space of symmetric positive definite (SPD) matrices.  
The main challenge is how to parametrize those distributions.   
Towards this end, we develop methods for embedding Gaussian distributions in matrix groups based on the theory of Lie groups and Riemannian symmetric spaces.   
Our method relies on the Riemannian geometry of the underlying manifolds and has the advantage of encoding high-order statistics from
3D joint positions.  
We show that the proposed method achieves competitive results in two-person interaction recognition on three benchmarks for 3D human activity understanding.   
\end{abstract}

\section{Introduction}
\label{sec:intro}

3D skeleton-based action recognition has been an active research topic in recent years 
with many potential applications. 
In this work, we focus on {\bf 3D skeleton-based two-person interaction recognition (3DTPIR)}. 
Compared to a large number of general-purpose methods for 3D skeleton-based action recognition, methods for 3DTPIR are much less studied in the literature. 
Recent works~\cite{perez2019interaction,YangPairwiseICIP20} have shown that state-of-the-art action recognition methods 
do not always perform well on human interaction recognition. 
This is because they lack an effective mechanism for capturing intra-person and inter-person 
joint relationships~\cite{perez2019interaction}.  

In two-person interactions, arm and leg movements are highly correlated. 
However, these correlations are simply encoded by distances between joints in most existing works for 3DTPIR~\cite{JiContrastFeat15,JiContrastInteract14,OUYEDMKL20,SBU-dataset12}.  
This motivates us to use high-order statistics, i.e. covariance matrices to better capture these interactions. 
It has been known that $n \times n$ covariance matrices lie on a special type of Riemannian manifolds, 
i.e. SPD manifolds (denoted by $Sym_n^{+}$).      
A large body of works has been developed for classification of SPD-valued data. 
Recently, SPD neural networks have demonstrated impressive results~\cite{HuangGool17}. 
One of the core issues that remains open is the finding of effective and efficient methods
for modeling probability distributions on $Sym_n^{+}$. 
Since Gaussian distributions (abbreviated as Gaussians) on $\mathbb{R}^n$ 
are the most popular probability distributions used in statistics, 
existing works mainly focused on generalizing them to $Sym_n^{+}$.  
Such a generalization was first given in~\cite{pennec:inria-00071490} in a more general context of Riemannian manifolds. 
However, the asymptotic formulae of Riemannian Gaussian distributions (abbreviated as Riemannian Gaussians) proposed in 
this work make them hard to evaluate and apply in practice.   
Some works aim to address this shortcoming by introducing notions of Riemannian Gaussians in 
symmetric spaces~\cite{SaidSPDGaussian17,SaidSymmetricGaussian18} and homogeneous spaces~\cite{chakraborty2017statistics}. 
These have been successfully applied to classification problems. 
In this work, we also interested in Riemannian Gaussians for classification.  
However, differently from the above works, we seek methods for embedding Riemannian Gaussians 
in matrix groups. This allows us to perform classification of Riemannian Gaussians
without having to resort to an exact expression of their probability density function as in~\cite{chakraborty2017statistics,SaidSPDGaussian17,SaidSymmetricGaussian18}.  

In summary, the main contributions of this work are: 
\begin{itemize}
\item We propose an embedding method for Gaussians by mapping them diffeomorphically to Riemannian symmetric spaces. 
\item We consider representing a 3D skeleton sequence by a set of SPD matrices that leads us to the study
of statistics on $Sym_n^{+}$. We show that the product space of mean and covariance on $Sym_n^{+}$ can be viewed 
as a Lie group with an appropriate group product. Moreover, we point out a connection between this space
and the group of lower triangular matrices with positive diagonal entries.   
\item Based on the theory described above, we introduce a neural network for learning a geometric representation from a 3D skeleton sequence.  
\item Experiments on three benchmarks for 3D human activity understanding demonstrate the competitiveness of our method with state-of-the-art methods. 
\end{itemize}

\section{Related Works}\label{sec:related_work}
We will briefly discuss representative works for 3DTPIR 
(Section~\ref{subsec:3d_action_recognition}), embeddings of Gaussians (Section~\ref{subsec:embedding_gaussian}), 
and probability distributions on $Sym_n^{+}$ (Section~\ref{subsec:statistics_spd}). 

\subsection{Two-person Interaction Recognition from 3D Skeleton Sequences}
\label{subsec:3d_action_recognition}  

A variety of approaches has been proposed for 3D skeleton-based action recognition.  
These are based on hand-crafted features~\cite{SkeQuad14,Luo13,SmedtCVPRW16,ActionletEns12,EigenJoints,ZhanGramMatrix16}
and deep learning~\cite{NN15,KeNewReRecCVPR17,LiuLSTM17,LiuTrustGateECCV16,LiuCVPR17LSTM,LIUEnhViewPR2017,LiuEvoMapsCVPR18,Nez2018,Shahroudy16NTU,Wang2017ModelingTD,
WengTraversalConvECCV18,ZhuCo-occu16}. Recent works focus on neural networks on manifolds~\cite{HuangGool17,Huang17DLLieGroup,HuangAAAI18,NguyenHandRecgCVPR19} and on graphs~\cite{ChenShiftGCN20,LiGraphConvAAAI18,LiActGCN19,SiEnhanceGCN19,YanAAAI18}. 
Due to space limit, we refer the interested reader to~\cite{ren2020survey} for a more comprehensive survey. 
Below we focus our discussion on 3DTPIR.  

Approaches for 3DTPIR are much less studied. 
Hand-crafted feature based methods mainly rely on distances~\cite{JiContrastFeat15,JiContrastInteract14,OUYEDMKL20,SBU-dataset12} 
or moving similarity~\cite{LiuStructuredNeuro18} between joints of two persons. 
Li and Leung~\cite{LiMultiviewInteract16} applied a multiple kernel learning
method to an interaction graph constructed from the relative variance of joint relative distances.    
Two-stream RNNs are proposed in~\cite{MenICPR20,Wang2017ModelingTD} where interactions between two persons are modeled 
by concatenating the 3D coordinates of their corresponding joints, or by augmenting
the input sequence with distances between their joints.  
In~\cite{perez2019interaction}, Relational Network~\cite{NIPS2017_e6acf4b0} is extended to automatically infer 
intra-person and inter-person joint relationships.  
The recent work~\cite{YangPairwiseICIP20} deals with graph construction in graph convolutional networks 
for 3DTPIR. 

\subsection{Embedding of Gaussians}
\label{subsec:embedding_gaussian}

Methods for embedding Gaussians are widely used in statistics, e.g. for measuring the distance between
probability distributions. 
The work of~\cite{Rao1992} first proposed a distance function based on the Fisher information as a Riemannian metric. 
However, in the general case of multivariate Gaussians,      
an exact formula for the distance function is difficult to obtain.     
In computer vision, one of the most widely used embedding is derived from~\cite{Lovric00}. 
The key idea is to identify Gaussians with SPD matrices by parametrizing the space of Gaussians as 
a Riemannian symmetric space. The work of~\cite{CALVO1990223} shares a similar idea of identifying Gaussians with SPD matrices.  
However, it is based on embedding Gaussians into the Siegel group. 
In~\cite{GongCVPR09}, a connection is established between Gaussians and a subspace of affine matrices.   
The method of~\cite{Li17} relies on the Log-Euclidean metrics~\cite{arsigny:inria-00070423} for embedding Gaussians in linear spaces.   

\subsection{Probability Distributions on $Sym_n^{+}$}
\label{subsec:statistics_spd}

Existing works mainly focused on generalizing Gaussians to $Sym_n^{+}$ 
due to their popularity in statistics.  
Generalizations of Gaussians are proposed in Riemannian manifolds~\cite{pennec:inria-00071490,ZhangPPGA13},  
symmetric spaces~\cite{SaidSPDGaussian17,SaidSymmetricGaussian18}, and homogeneous spaces~\cite{chakraborty2017statistics}. 
In~\cite{Barbaresco2019,BrooksRieBatNorm19}, Riemannian Gaussians are derived from the definition 
of maximum entropy on exponential families. 
Family of Alpha-Divergences and other related divergences have also been extensively studied~\cite{CichockiABG10}. 

\section{Background Theory}
\label{sec:background}

\subsection{Statistics on Riemannian Manifolds}
\label{subsec:statistics_background}
  
The theory presented here is based on~\cite{pennec:inria-00071490}. 
In this framework, the structure of a manifold $\mathcal{M}$ is specified by a Riemannian metric. 
Let $x$ be a point of $\mathcal{M}$ as a local reference and $T_x \mathcal{M}$ be the tangent space at $x$. 
A Riemannian metric is a continuous collection of dot products $<.|.>_x$ on $T_x\mathcal{M}$.
The distance between two points of a connected Riemannian manifold is the minimum length among the smooth curves joining them.  
The curves realizing this minimum for any two points are called geodesics. 
Let $v \in T_x \mathcal{M}$ be a tangent vector at $x$.   
We define the exponential map at $x$ as the function that maps $v$ 
to the point $y \in \mathcal{M}$ that is reached after a unit time by the geodesic starting at $x$ with this tangent vector. 
This map is defined in the whole tangent space $T_x \mathcal{M}$ but it is generally one-to-one only locally around 0
in the tangent space (i.e. around $x$ in the manifold). 
Let 
$\overrightarrow{xy} = \log_x(y)$ be the inverse of the exponential map that is the smallest vector 
as measured by the Riemannian metric such that $y = \exp_x(\overrightarrow{xy})$.  
The exponential chart at $x$ can be seen as the development of $\mathcal{M}$ in the tangent space at a given point along the geodesics. 

The definitions of mean and covariance on a Riemannian manifold are given below. 
\begin{definition}
Let $\mathbf{x}$ be a random point of probability density function $p_{\mathbf{x}}$.   
Denote by $\text{dist}_R(y,x)$ the distance between $x,y \in \mathcal{M}$ induced by the Riemannian
metric of $\mathcal{M}$. The set of expected or mean values is: 
\begin{equation}\label{eq:riemannian_mean}
\mathbb{E}[\mathbf{x}] = \argmin_{y \in \mathcal{M}} \Big( \int_{\mathcal{M}} \text{dist}_R(y,z)^2.p_{\mathbf{x}}(z).d\mathcal{M}(z) \Big),
\end{equation} 
where $d\mathcal{M}(z)$ is the volume measure induced by the Riemannian metric of $\mathcal{M}$, 
and we assume that the integral is finite for all points $y \in \mathcal{M}$ 
(which is true for a density with a compact support). 
\end{definition}

\begin{definition}
Let $\mathbf{x}$ be a random point, $\bar{x}$ be a mean value that we assume to be
unique. The covariance is defined as: 
\begin{equation}\label{eq:riemannian_cov}
\text{Cov}_{\bar{x}}(\mathbf{x}) = \mathbf{E}[\overrightarrow{\bar{x}\mathbf{x}}.\overrightarrow{\bar{x}\mathbf{x}}^T] = \int_{\mathcal{D}(\bar{x})} (\overrightarrow{\bar{x}\mathbf{x}}).(\overrightarrow{\bar{x}\mathbf{x}})^T.p_{\mathbf{x}}(x).d\mathcal{M}(x),
\end{equation}
where $\mathcal{D}(\bar{x})$ is the maximal definition domain for the exponential chart at $\bar{x}$. 
\end{definition}

\section{Proposed Approach}
\label{sec:proposed_approach}

In what follows, we are interested in matrices over the field of real numbers, unless otherwise stated. 

\subsection{Embedding of Gaussians}
\label{subsec:embedding_riemannian_symmetric_space}

Let $N(n)$ be the space of $n-$variate Gaussians, $Sym^{+,1}_{n}$ be the space of SPD $n \times n$ matrices with determinant 1. 
Lemma~\ref{lem_gauss_embed} gives an embedding of Gaussians in $N(n)$. 
\begin{lemma}\label{lem_gauss_embed}
We can identify the Gaussian $(\pmb{\Sigma},\pmb{\mu}) \in N(n)$ with 
the following matrix in $Sym^{+,1}_{n+k}$:  
\begin{equation}\label{eq:embed_final}
(\det \pmb{\Sigma})^{-\frac{1}{n+k}} \begin{bmatrix} \pmb{\Sigma} + k\pmb{\mu} \pmb{\mu}^T & \pmb{\mu}(k) \\ \pmb{\mu}(k)^T & \mathbf{I}_k \end{bmatrix},
\end{equation} 
where $\pmb{\mu}$ and $\pmb{\Sigma}$ are the mean and covariance of the Gaussian, 
$\mathbf{I}_k$ is the $k \times k$ identity matrix, 
$\pmb{\mu}(k)$ is a matrix with $k$ identical column vectors $\pmb{\mu}$. 
\end{lemma}

{\bf Proof.} See supplementary material. 

When $k=1$, the embedding in~(\ref{eq:embed_final}) becomes the one introduced in~\cite{Lovric00}. 
The natural symmetric Riemannian metric resulting from the above embedding is given in Lemma~\ref{lem_riemannian_metric}.    
\begin{lemma}\label{lem_riemannian_metric} 
The Riemannian metric is given by:
\begin{align}\label{eq:riemannian_metric}
\begin{split}
<\mathbf{A}_1,\mathbf{A}_2>_{\mathbf{P}} = &Tr(\mathbf{A}_1\mathbf{P}^{-1}\mathbf{A}_2\mathbf{P}^{-1}) - \\ & - \frac{1}{n+k}Tr(\mathbf{A}_1\mathbf{P}^{-1})Tr(\mathbf{A}_2\mathbf{P}^{-1}), 
\end{split}
\end{align}
\end{lemma}
where $\mathbf{A}_1$ and $\mathbf{A}_2$ are two tangent vectors at $\mathbf{P}$.   

{\bf Proof.} See supplementary material. 

It turns out that the Riemannian metric given in~(\ref{eq:riemannian_metric}) belongs to the family of 
affine-invariant metrics proposed in~\cite{pennec:tel-00633163}.  
Consequently, the exponential map at a point can be obtained~\cite{pennec:inria-00070743} as:
\begin{equation}\label{eq:exponential_map_airm} 
\exp_{\mathbf{P}}(\mathbf{A}) = \mathbf{P}^{\frac{1}{2}} \exp \Big( \mathbf{P}^{-\frac{1}{2}} \mathbf{A} \mathbf{P}^{-\frac{1}{2}} \Big) \mathbf{P}^{\frac{1}{2}},
\end{equation}  
where $\mathbf{P}$ is a SPD matrix, $\mathbf{A}$ is a tangent vector at $\mathbf{P}$, and $\exp(.)$ is the matrix exponential. By inverting the exponential map, we obtain the logarithmic map:
\begin{equation}\label{eq:logarithmic_map_airm} 
\log_{\mathbf{P}}(\mathbf{Q}) = \mathbf{P}^{\frac{1}{2}} \log \Big( \mathbf{P}^{-\frac{1}{2}} \mathbf{Q} \mathbf{P}^{-\frac{1}{2}} \Big) \mathbf{P}^{\frac{1}{2}},  
\end{equation} 
where $\mathbf{P}$ and $\mathbf{Q}$ are two SPD matrices, and $\log(.)$ is the matrix logarithm.   

\subsection{Statistics on $Sym_n^{+}$}
\label{subsec:parallel_transport}

Suppose that we are given a set of matrices $\mathbf{P}_1,\mathbf{P}_2,\ldots,\mathbf{P}_L \in Sym_n^{+}$.  
From~(\ref{eq:riemannian_mean}), one can define the empirical or discrete mean value 
of $\mathbf{P}_1,\mathbf{P}_2,\ldots,\mathbf{P}_L$ as:
\begin{equation}
\mathbf{P}^m = \argmin_{y \in Sym_{n}^{+}} \Big( \frac{1}{L}\sum_{i=1}^L \text{dist}_R(y,\mathbf{P}_i)^2 \Big).  
\end{equation}

The mean can be computed by an iterative procedure consisting in: (1) projecting the SPD matrices in the tangent space at the current mean; 
(2) estimating the arithmetic mean in that space; (3) projecting the mean back in $Sym^{+}_n$.  
These steps are iterated until convergence~\cite{Moakher_adifferential}.    

Similarly, the empirical covariance of a set of $L$ SPD matrices of mean $\mathbf{P}^m$ is defined using the discrete version of the expectation operator in~(\ref{eq:riemannian_cov}): 
\begin{equation}
\mathbf{P}^c = \frac{1}{L-1} \sum_{i=1}^L \overrightarrow{\mathbf{P}^m\mathbf{P}_i} \otimes \overrightarrow{\mathbf{P}^m\mathbf{P}_i},
\end{equation}
where $\otimes$ denotes the tensor product. 
 
We propose to learn a transformation of $\overrightarrow{\mathbf{P}^m\mathbf{P}_i}$ from 
$T_{\mathbf{P}^m}Sym^{+}_{n}$ to another tangent space
so that the covariance computed in this space is more discriminative for classification. 
The transformation is performed by parallel transport (PT). 
We need Lemma~\ref{lem_pt} for our transformation.  
\begin{lemma}\label{lem_pt} 
Let $\mathbf{P},\mathbf{Q} \in Sym^{+}_{n}$. Let the Riemannian metric be the one given in~(\ref{eq:riemannian_metric}). 
The PT from $\mathbf{Q}$ to $\mathbf{P}$ along geodesics connecting $\mathbf{Q}$ and $\mathbf{P}$ 
of a tangent vector $\mathbf{A} \in T_{\mathbf{Q}}Sym^{+}_{n}$ is given by:
\begin{equation}\label{eq:parallel_transport}
\mathcal{T}_{\mathbf{Q},\mathbf{P}}(\mathbf{A}) \triangleq (\mathbf{P}\mathbf{Q}^{-1})^{\frac{1}{2}}\mathbf{A}\big((\mathbf{P}\mathbf{Q}^{-1})^{\frac{1}{2}}\big)^T. 
\end{equation}
\end{lemma}

{\bf Proof.} See supplementary material. 

The formula of PT in~(\ref{eq:parallel_transport}) is the same as those  
in~\cite{FerreiraRie06,SraSIAM15,YairPT19} which are all based on the 
Riemannian metric of the following form:   
\begin{equation}\label{eq:airm_metric}
<\mathbf{A}_1,\mathbf{A}_2>_{\mathbf{P}} = Tr(\mathbf{A}_1\mathbf{P}^{-1}\mathbf{A}_2\mathbf{P}^{-1}), 
\end{equation} 
where $\mathbf{A}_1$ and $\mathbf{A}_2$ are two tangent vectors at $\mathbf{P}$.   
 
In~\cite{BrooksRieBatNorm19}, the authors also use PT for designing Riemannian batch normalization (RBN) layers. 
Our method differs from theirs in three main aspects. First, their method learns the parameters of RBN layers 
from the statistics of mini-batches, while our method deals with the statistics within each sequence. 
Note that a RBN layer can also be designed in our framework and can potentially improve the accuracy of our network.   
Second, their formulation of Riemannian Gaussians involves only a Riemannian mean without notion of variance.      
Third, their method does not aim to leverage the second-order statistics (covariance) on SPD manifolds.  

Now suppose that $\overrightarrow{\mathbf{P}^m\mathbf{P}_i},i=1,\ldots,L$ are transported to another tangent space 
at $\tilde{\mathbf{P}}$ (the target point of PT), then 
the covariance can be estimated as: 
\begin{equation}
\mathbf{P}^c = \frac{1}{L-1} \sum_{i=1}^L \mathcal{T}_{\mathbf{P}^m,\tilde{\mathbf{P}}}(\overrightarrow{\mathbf{P}^m\mathbf{P}_i}) \otimes \mathcal{T}_{\mathbf{P}^m,\tilde{\mathbf{P}}}(\overrightarrow{\mathbf{P}^m\mathbf{P}_i}).
\end{equation}

Let $f_{v}(.)$ be a mapping that vectorizes a symmetric matrix by taking its lower triangular part
and applying a $\sqrt{2}$ coefficient on its off-diagonal entries in order to
preserve the norm~\cite{pennec:inria-00070743}. Then, the covariance is given by:    
\begin{equation}\label{eq:vectorize_cov}
\mathbf{P}^c = \frac{1}{L-1} \sum_{i=1}^L f_{v}\big(\mathcal{T}_{\mathbf{P}^m,\tilde{\mathbf{P}}}(\overrightarrow{\mathbf{P}^m\mathbf{P}_i})\big) f_{v}\big(\mathcal{T}_{\mathbf{P}^m,\tilde{\mathbf{P}}}(\overrightarrow{\mathbf{P}^m\mathbf{P}_i})\big)^T.
\end{equation}

If $\mathbf{P}^c$ is a matrix of size $n' \times n'$, then we use a point $(\mathbf{P}^m,\mathbf{P}^c)$ 
that lies on the product manifold $Sym_n^{+} \times Sym_{n'}^{+}$ 
to parametrize the distribution of the given set of SPD matrices. Next, 
we propose an embedding of this point based on the Lie group theory.   

\subsection{Embedding of Riemannian Gaussians}
\label{subsec:proposed_theory}
  
We first define an appropriate group product on the product manifold $Sym_n^{+} \times Sym_{n'}^{+}$.  

\begin{definition} 
Let $\mathcal{M}(n,n')$ be the product manifold $Sym_n^{+} \times Sym_{n'}^{+}$.    
Let $(\mathbf{P}^m_i, \mathbf{P}^c_i) \in \mathcal{M}(n,n'),i=1,2$  
where $\mathbf{P}^m_i \in Sym_n^{+}$, $\mathbf{P}^c_i \in Sym_{n'}^{+}$, and $\mathbf{P}^c_i=\mathbf{L}_i\mathbf{L}_i^T$ be
the Cholesky decomposition of $\mathbf{P}^c_i$. 
Denote by $\varphi: Sym_{n}^{+} \rightarrow M^{k' \times n'}$ a smooth bijective mapping with a smooth inverse where $M^{k' \times n'}$ 
is a subset of the set of $k' \times n'$ matrices.  
The group product $\star$ between two elements of $\mathcal{M}(n,n')$ is defined as:
\begin{align}\label{eq:multiplication}
\begin{split}
\star: & \mathcal{M}(n,n') \times \mathcal{M}(n,n') \rightarrow \mathcal{M}(n,n') \\ &
(\mathbf{P}^m_1,\mathbf{P}^c_1) \star (\mathbf{P}^m_2,\mathbf{P}^c_2) \\ & = (\varphi^{-1}(\varphi(\mathbf{P}^m_1) \mathbf{L}_2  + \varphi(\mathbf{P}^m_2)), (\mathbf{L}_1 \mathbf{L}_2)(\mathbf{L}_1 \mathbf{L}_2)^T). 
\end{split}
\end{align}
\end{definition}

Theorem~\ref{theorem_lie_group} shows that $\mathcal{M}(n,n')$ forms a Lie group. 

\begin{theorem}\label{theorem_lie_group} 
$\mathcal{M}(n,n')$ is a Lie group under product $\star$. 
\end{theorem}

{\bf Proof.} See supplementary material.  

Based on Theorem~\ref{theorem_lie_group}, we can establish a Lie group isomorphism between $\mathcal{M}(n,n')$ 
and a subgroup of a group of lower triangular matrices with positive diagonal entries. 

\begin{theorem} 
Denote by 
$LT^{+}(n')$ the group of lower triangular $n' \times n'$ matrices with positive diagonal entries, 
$\mathbf{0}_{n' \times k'}$ the $n' \times k'$ matrix with all elements equal to zero.  
Let
\begin{equation}
K^{+}(n'+k') = \left\{ \mathbf{K}_{\mathbf{P}^m,\mathbf{H}}  \triangleq \begin{bmatrix} \mathbf{H} & \mathbf{0}_{n' \times k'} \\ \varphi(\mathbf{P}^m) & \mathbf{I}_{k'} \end{bmatrix} \right\}, 
\end{equation}
where $\mathbf{H} \in LT^{+}(n')$, and 
\begin{equation}
\phi: K^{+}(n'+k') \rightarrow \mathcal{M}(n,n'), \phi(\mathbf{K}_{\mathbf{P}^m,\mathbf{L}}) = (\mathbf{P}^m,\mathbf{P}^c),
\end{equation}
where $\mathbf{P}^c = \mathbf{L}\mathbf{L}^T$, $\mathbf{L} \in LT^{+}(n')$. Then $\phi$ is a Lie group isomorphism. 
\end{theorem}

{\bf Proof.} See supplementary material. 

We now can give the embedding matrix of a point $(\mathbf{P}^m,\mathbf{P}^c) \in \mathcal{M}(n,n')$ 
where $\mathbf{P}^c = \mathbf{L}\mathbf{L}^T$ as follows:
\begin{equation}\label{eq:spdstats_embedding}
(\mathbf{P}^m,\mathbf{P}^c) \mapsto \begin{bmatrix} \mathbf{L} & \mathbf{0}_{n' \times k'} \\ \varphi(\mathbf{P}^m) & \mathbf{I}_{k'} \end{bmatrix}. 
\end{equation}

The embedding matrix in~(\ref{eq:spdstats_embedding}) depends on the choice of function $\varphi(.)$. 
In this work, we set $\varphi = (f_v \circ f_{lm})(k')^T$ where $(f_v \circ f_{lm})(k')^T$ is  
the transpose of $(f_v \circ f_{lm})(k')$, $(f_v \circ f_{lm})(k')$ is a matrix with $k'$ identical column vectors 
obtained from $f_v \circ f_{lm}$, and $f_{lm}(.)$ is given by: 
\begin{equation}
f_{lm}(\mathbf{P}) = \log(\mathbf{P}) = \mathbf{U} \log(\mathbf{Z}) \mathbf{U}^T,   
\end{equation}
where $\mathbf{P} = \mathbf{U} \mathbf{Z} \mathbf{U}^T$ is the eigenvalue decomposition of $\mathbf{P}$, 
and $\log(\mathbf{Z})$ is the diagonal matrix of eigenvalue logarithms. 

\subsection{A Neural Network for 3DTPIR} 
\label{subsec:proposed_network}

\begin{figure}[t]
  \begin{center}
    \begin{tabular}{c}      
      \includegraphics[width=0.7\linewidth, clip=true]{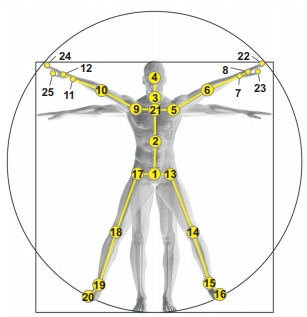}
    \end{tabular}
  \end{center} 
  \caption{\label{fig:body_skeletal} Illustration of body joints' positions (only the first skeleton is shown). The joint at the hip is selected as the root of the skeleton. The distance between joints 1 and 6 is 4. 
The joints 22,23,24,25 are not used in the convolution operation. 
The arms then contain the following joints: $5,6,7,8,9,10,11,12$. 
The legs contain the following joints: $13,14,15,16,17,18,19,20$ 
(figure reproduced from~\cite{Shahroudy16NTU}).}    
\end{figure}

We are now ready to introduce a neural network (GeomNet) for 3DTPIR based on the theory developed in the previous sections.   
Let $N^j$ and $N^f$ be the number of joints and that of frames in a given sequence, respectively,
Let $\mathbf{x}^{in}_{t,i} \in \mathbb{R}^3,t=1,\ldots,N^f,i=1,\ldots,N^j$ be the feature vector (3D coordinates)
of joint $i$ at frame $t$.  
Two joints $i$ and $j$ are neighbors if they are connected by a bone. 
Denote by $\mathcal{S}_i$ the set of neighbors of joint $i$. 
Let $i^{1,r},i^{2,r}$ be the two joints selected as the roots of the first and second skeleton, 
respectively (see Fig.~\ref{fig:body_skeletal}). 
For any two joints $i$ and $i^r \in \{ i^{1,r},i^{2,r} \}$ that belong to the same skeleton, the distance $\text{dist}_J(i,i^r)$ 
between them is defined as the number of bones connecting them (see Fig.~\ref{fig:body_skeletal}).  
The first layer of GeomNet is a convolutional layer written as:        
\begin{equation}\label{eq:graph_conv}
\mathbf{x}^{out}_{t,i} = \sum_{t'=t-1}^{t+1} \sum_{i' \in \mathcal{S}_i} \tilde{\mathbf{W}}_{t',i'} \mathbf{x}^{in}_{t',i'},
\end{equation}
where $\mathbf{x}^{out}_{t,i} \in \mathbb{R}^{d}$ is the output feature vector of joint $i$ at frame $t$, 
and $\tilde{\mathbf{W}}_{t',i'}$ is defined as:
\begin{equation}
\tilde{\mathbf{W}}_{t',i'} = \begin{cases} \mathbf{W}_{t'+2-t,1}, & \text{if dist}_J(i',i^r) = \text{dist}_J(i,i^r)-1 \\ \mathbf{W}_{t'+2-t,2}, & \text{if }i'=i  \\ \mathbf{W}_{t'+2-t,3}, & \text{if dist}_J(i',i^r) = \text{dist}_J(i,i^r)+1 \end{cases}
\end{equation}

Here, the set of weights $\{ \mathbf{W}_{u,v} \},u,v=1,2,3$ completely defines the convolution filters in Eq.~(\ref{eq:graph_conv}).   
Let $N^{j,1}$ and $N^{j,2}$ be the numbers of joints belonging to the arms and legs 
of two skeletons, respectively (see Fig.~\ref{fig:body_skeletal}).  
Let $\mathbf{X}^{out,1}$ and $\mathbf{X}^{out,2}$ respectively of size $N^{j,1} \times N^f \times d$ and $N^{j,2} \times N^f \times d$  
be the data associated with the arms and legs of two persons. 
The motivation behind this partition is that the interaction between two persons often involve those among their arms and
those among their legs. 
For each $b \in \{ 1,2 \}$, the set of $N^{j,b}N^f$ $d-$dim feature vectors from $\mathbf{X}^{out,b}$ 
is partitioned into $L$ subsets using K-means clustering.   
Let $\mathbf{y}^b_{l,1},\ldots,\mathbf{y}^b_{l,i^b_l}$ be the feature vectors in the $l^{th}$ subset.  
We assume that $\mathbf{y}^b_{l,1},\ldots,\mathbf{y}^b_{l,i^b_l}$ are i.i.d. samples from a Gaussian
$(\pmb{\Sigma}^b_l,\pmb{\mu}^b_l)$ whose parameters can be estimated as:  

\begin{equation}\label{eq:mean}
\pmb{\mu}^b_l = \frac{1}{i^b_l} \sum_{j=1}^{i^b_l} \mathbf{y}^b_{l,j},
\end{equation}
\begin{equation}\label{eq:cov}
\pmb{\Sigma}^b_l = \frac{1}{i^b_l-1} \sum_{j=1}^{i^b_l} (\mathbf{y}^b_{l,j} - \pmb{\mu}^b_l)(\mathbf{y}^b_{l,j} - \pmb{\mu}_l^b)^T.
\end{equation}
  
Based on the theory developed in Section~\ref{subsec:embedding_riemannian_symmetric_space}, the Gaussian
$(\pmb{\Sigma}^b_l,\pmb{\mu}^b_l)$ can be identified with the following matrix:
\begin{equation}\label{eq:spd}
\mathbf{P}^b_l = (\det \pmb{\Sigma}_l^b)^{-\frac{1}{n+k}} \begin{bmatrix} \pmb{\Sigma}^b_l + k\pmb{\mu}^b_l (\pmb{\mu}_l^b)^T & \pmb{\mu}^b_l(k) \\ (\pmb{\mu}_l^b(k))^T & \mathbf{I}_k \end{bmatrix}.
\end{equation}

The above computations can be performed by a layer as:
\begin{equation}
\{ \mathbf{P}^b_l \}^{b=1,2}_{l=1,\ldots,L} = f_{gaussemb}(\{ \mathbf{X}^{out,b} \}_{b=1,2}).
\end{equation}

The next layer is designed to compute statistics on SPD manifolds and can be written by:
\begin{equation}
\{ \mathbf{P}^{b,m},\mathbf{P}^{b,c} \}_{b=1,2} = f_{spdstats}\big(\{ \mathbf{P}^b_l,\mathbf{W}^b_{pt} \}^{b=1,2}_{l=1,\ldots,L}\big),
\end{equation}
where $\mathbf{W}^b_{pt},b=1,2$ are the parameters corresponding to the target points of PT (see Section~\ref{subsec:parallel_transport}). 
Specifically, $\mathbf{P}^{b,m}$ is the mean of $\mathbf{P}^b_l,l=1,\ldots,L$, and $\mathbf{P}^{b,c}$ is given by:  
\begin{align}
\begin{split}
\mathbf{P}^{b,c} = \frac{1}{L-1} \sum_{i=1}^L & f_{v}\big(\mathcal{T}_{\mathbf{P}^{b,m},\mathbf{W}^b_{pt}}(\overrightarrow{\mathbf{P}^{b,m}\mathbf{P}_i})\big) \times \\ & f_{v}\big(\mathcal{T}_{\mathbf{P}^{b,m},\mathbf{W}^b_{pt}}(\overrightarrow{\mathbf{P}^{b,m}\mathbf{P}_i})\big)^T.
\end{split}
\end{align} 

The next layer computes the embeddings of statistics $\mathbf{P}^{b,m},\mathbf{P}^{b,c},b=1,2$ 
and can be written as: 
\begin{equation}\label{eq:layer_spdstatsemb}
\{ \mathbf{B}^b \}_{b=1,2} = f_{spdstatsemb}\big(\{ \mathbf{P}^{b,m},\mathbf{P}^{b,c} \}_{b=1,2}\big),
\end{equation}
where $\mathbf{B}^b$ is the embedding matrix of $(\mathbf{P}^{b,m},\mathbf{P}^{b,c})$ given in the right-hand side of~(\ref{eq:spdstats_embedding}). 

The next layer transforms $\mathbf{B}^b,b=1,2$ to some matrices in $LT^{+}(n'+k')$ as:
\begin{equation}
\{ \mathbf{D}^b \}_{b=1,2} = f_{trilmap}\big( \{ \mathbf{B}^b,\mathbf{W}^b_{lw} \}_{b=1,2} \big),
\end{equation}
where $\mathbf{D}^b = \mathbf{B}^b \mathbf{W}^b_{lw}$, $\mathbf{W}^b_{lw},b=1,2$ are the parameters
that are required to be in $LT^{+}(n'+k')$ so that the outputs $\mathbf{D}^b$ are also in $LT^{+}(n'+k')$.  
The network then performs a projection:  
\begin{equation}
\{ \mathbf{E}^b \}_{b=1,2} = f_{triltoeud}(\{ \mathbf{D}^b \}_{b=1,2}),
\end{equation} 
where $\mathbf{E}^b = f_{lm}(\mathbf{D}^b(\mathbf{D}^b)^T),b=1,2$.   
Finally, a fully-connected (FC) layer and a softmax layer are used to obtain class probabilities:
\begin{equation}
\mathbf{C}^{out} = f_{prob}\big(concat(f_v(\mathbf{E}^1),f_v(\mathbf{E}^2)), \mathbf{W}_{fc}\big),
\end{equation}
where $\mathbf{W}_{fc}$ are the parameters of the FC layer, the operator $concat(\mathbf{V}_1,\mathbf{V}_2)$
concatenates the two column vectors $\mathbf{V}_1$ and $\mathbf{V}_2$ vertically, 
and $\mathbf{C}^{out}$ are the output class probabilities. 
We use the cross-entropy loss for training GeomNet.        

\subsection{Geometry Aware Constrained Optimization}
\label{subsec:geom_optimization} 

Some layers of GeomNet rely on the eigenvalue decomposition. 
To derive the backpropagation updates for these layers,  
we follow the framework of~\cite{Ionescu2015} for computation of the involved partial derivatives. 
The optimization procedure for the parameters $\mathbf{W}^b_{pt},\mathbf{W}^b_{lw},b=1,2$ 
is based on the Adam algorithm in Riemannian manifolds~\cite{BecGan19}.  
The Riemannian Adam update rule is given by: 
\begin{equation}\label{eq:radam_update_rule}
x_{t+1} = \exp_{x_t} \Bigg(- \alpha \frac{\hat{m}_t}{\sqrt{\hat{v}_t} + \epsilon} \Bigg),    
\end{equation}   
where $x_t$ and $x_{t+1}$ are respectively the parameters updated at timesteps $t$ and $t+1$, 
$\hat{m}_t = m_t/(1-\beta_1^t)$, $\hat{v}_t = v_t/(1-\beta_2^t)$,  
$m_t = \beta_1 \tau_{t-1} + (1 - \beta_1)g_t$ is a momentum term,  
$v_t = \beta_2 v_{t-1} + (1 - \beta_2)\|g_t\|^2_{x_t}$ is an adaptivity term, 
$g_t$ is the gradient evaluated at timestep $t$, 
$\alpha,\epsilon,\beta_1,\beta_2$ are constant values.   
The squared Riemannian norm $\|g_t\|^2_{x_t}=<g_t|g_t>_{x_t}$ corresponds to the squared gradient value 
in Riemannian settings. Here, $<.|.>_{x_t}$ is the dot product for the Riemannian metric of the 
manifold in consideration, as discussed in Section~\ref{subsec:statistics_background}. 
After updating $x_{t+1}$ in Eq.~(\ref{eq:radam_update_rule}), 
we update $\tau_t$ as the PT of $m_t$ along geodesics connecting $x_t$ and $x_{t+1}$, 
i.e. $\tau_t = \mathcal{T}_{x_t,x_{t+1}}(m_t)$.  
  
The update rule in Eq.~(\ref{eq:radam_update_rule}) requires the computation of the exponential map 
and the PT. 
For SPD manifolds, these operations are given in Eqs.~(\ref{eq:exponential_map_airm}) and~(\ref{eq:parallel_transport}).   
It remains to define these operations for the update of the parameters $\mathbf{W}^b_{lw},b=1,2$. 
To this aim, we rely on the Riemannian geometry of $LT^{+}(n)$ studied in the recent work~\cite{Lin_2019}.   
By considering the following metric: 
\begin{equation} 
<\mathbf{U},\mathbf{V}>_{\mathbf{K}} = \sum_{i > j} \mathbf{U}_{ij}\mathbf{V}_{ij} + \sum_{j=1}^n \mathbf{U}_{jj} \mathbf{V}_{jj} \mathbf{K}_{jj}^{-2}, 
\end{equation} 
where $\mathbf{K} \in LT^{+}(n)$, $\mathbf{U},\mathbf{V} \in T_{\mathbf{K}}LT^{+}(n)$,    
$\mathbf{U}_{ij}$ is the element on the $i^{th}$ row and $j^{th}$ column of $\mathbf{U}$, 
Lin has shown~\cite{Lin_2019} that the space $LT^{+}(n)$ (referred to as Cholesky space) 
equipped with the above metric forms a Riemannian manifold.  
On this manifold, the exponential map at a point can be computed as:  
\begin{equation}
\exp_{\mathbf{K}} \mathbf{U} = \lfloor \mathbf{K} \rfloor + \lfloor \mathbf{U} \rfloor + \mathbb{D}(\mathbf{K})\exp ( \mathbb{D}(\mathbf{U})\mathbb{D}(\mathbf{K})^{-1} ), 
\end{equation}  
where $\mathbf{K} \in LT^{+}(n)$, $\mathbf{U} \in T_{\mathbf{K}}LT^{+}(n)$, $\lfloor \mathbf{K} \rfloor$ is a matrix of the same size 
as $\mathbf{K}$ whose $(i,j)$ element is $\mathbf{K}_{ij}$ if $i > j$ and is zero otherwise, 
$\mathbb{D}(\mathbf{K})$ is a diagonal matrix whose $(i,i)$ element is $\mathbf{K}_{ii}$. 
Also, the PT of a tangent vector $\mathbf{U} \in T_{\mathbf{K}}LT^{+}(n)$ 
to a tangent vector at $\mathbf{H} \in LT^{+}(n)$ is given by:
\begin{equation}
\mathcal{T}_{\mathbf{K},\mathbf{H}}(\mathbf{U}) = \lfloor \mathbf{U} \rfloor + \mathbb{D}(\mathbf{H}) \mathbb{D}(\mathbf{K})^{-1} \mathbb{D}(\mathbf{U}),  
\end{equation}     
where $\mathbf{K},\mathbf{H} \in LT^{+}(n)$. 

\section{Experiments}
\label{sec:exps} 
 
Our network was implemented with Tensorflow deep learning framework and the experiments were conducted using two NVIDIA GeForce GTX 1080 GPUs. 
We used GeomStats library~\cite{MiolaneGeomstats18} for geometric computations.  
The dimension $d$ of output vectors at the convolutional layer, the number of clusters $L$, 
and the learning rate were set to 9, 180, and $10^{-2}$, respectively. 
The batch sizes were set respectively to 30 and 256 for the experiments on SBU Interaction dataset and those on NTU datasets.   
The values of the pair $(k,k')$ (see~(\ref{eq:embed_final}) and~(\ref{eq:spdstats_embedding})) were set to $(2,3)$ and $(2,1)$  
for the experiments on SBU Interaction and NTU datasets, respectively. 
The values of $\alpha,\epsilon,\beta_1$, and $\beta_2$ 
in the Riemannian Adam algorithm\footnote{Our code deals with constrained and unconstrained parameters.} 
were set to $10^{-3},10^{-8},0.9$, and $0.999$, respectively~\cite{KingmaICLR19}.   
In our experiments, GeomNet converged well after 600 epochs. 
For more details on our experiments, we refer the interested reader to the supplementary material. 

\subsection{Datasets and Experimental Settings}
\label{subsec:human_interaction_recognition}

{\bf SBU Interaction dataset.} 
This dataset~\cite{SBU-dataset12} contains 282 sequences in 8 action classes created from 7 subjects. 
Each action is performed by two subjects 
where each subject has 15 joints. The joints 4,21,1,5,6,7,9,10,11,13,14,15,17,18,19 in Fig.~\ref{fig:body_skeletal}
correspond respectively to the joints 1,2,3,4,5,6,7,8,9,10,11,12,13,14,15 of the first skeleton of SBU Interaction dataset.   
We followed the experimental protocol based on 5-fold cross validation with 
the provided training/testing splits~\cite{SBU-dataset12}.  

{\bf NTU RGB+D 60 dataset.}
This dataset~\cite{Shahroudy16NTU} contains 56,880 sequences 
created from 40 subjects with three cameras views and categorized into 60 classes.  
We followed the two experimental protocols cross-subject (X-subject) and cross-view (X-view)~\cite{Shahroudy16NTU}. 

{\bf NTU RGB+D 120 dataset.}
This dataset~\cite{Liu_2019_NTURGBD120}  
contains 114,480 sequences in 120 action classes, captured 
by 106 subjects with three cameras views. 
We followed the two experimental protocols cross-subject (X-subject) and cross-setup (X-setup)~\cite{Liu_2019_NTURGBD120}. 

\subsection{Ablation Study}
\label{subsec:ablation_study}

In this section, we study the impact of different components of GeomNet on its accuracy\footnote{The results of GeomNet are averaged over 3 runs.} on SBU Interaction and NTU RGB+D 60 datasets.  

\begin{figure}[t]
\centering
\includegraphics[width=0.5\textwidth, trim = 0 60 0 100, clip=true]{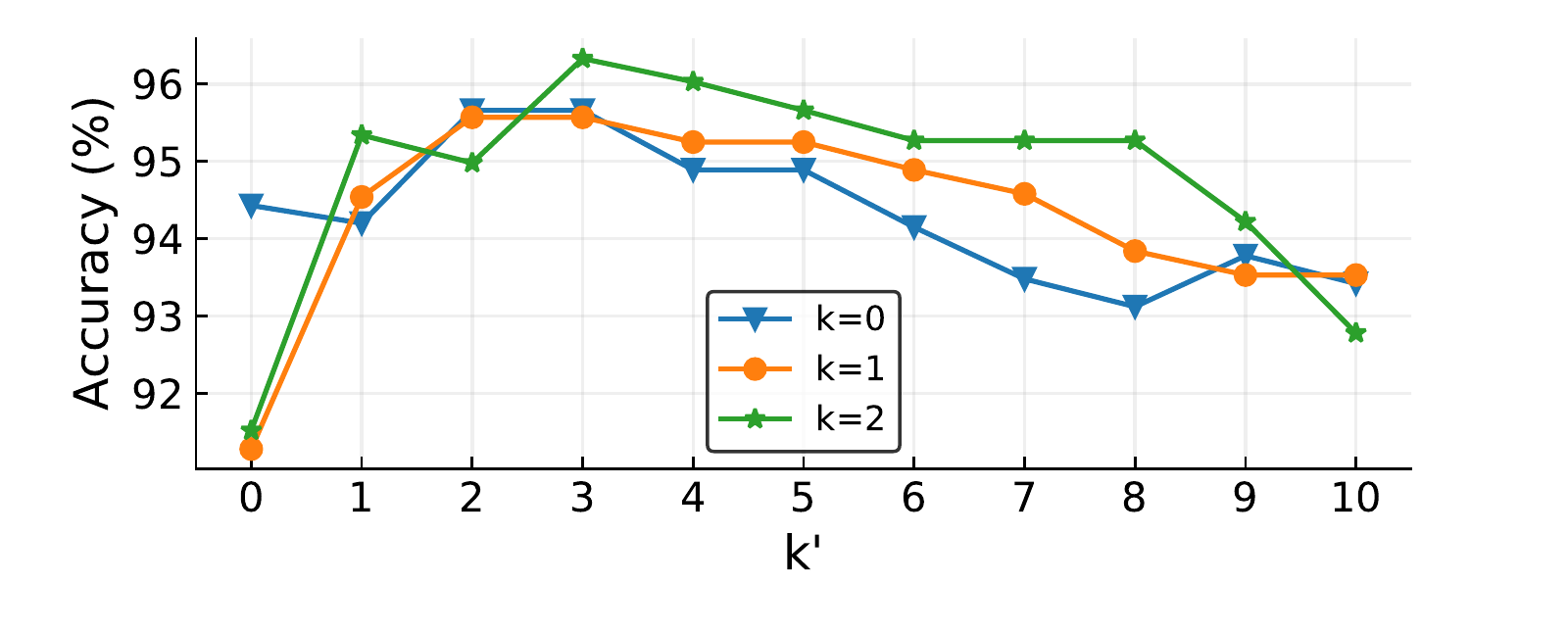}
\caption{\label{fig_a:accuracy_k1k2} Accuracy of GeomNet on SBU Interaction dataset with different settings of $(k,k')$.}  
\end{figure}

{\bf Embedding dimensions.} Here we investigate the impact of the parameters $k$ and $k'$  
(see~(\ref{eq:embed_final}) and~(\ref{eq:spdstats_embedding})). 
Fig.~\ref{fig_a:accuracy_k1k2} shows the accuracies of GeomNet on SBU Interaction dataset with different settings of $(k,k')$,  
i.e. $k=0,1,2$ and $k'=0,\ldots,10$. Note that when $k=0$, the layer $f_{gaussemb}$ relies only on the 
covariance information. Also, when $k'=0$, the outputs $\mathbf{B}^b,b=1,2$ of the layer $f_{spdstatsemb}$ 
are simply obtained by the Cholesky decomposition of $\mathbf{P}^{b,c}$, i.e. $\mathbf{B}^b(\mathbf{B}^b)^T = \mathbf{P}^{b,c}$.  
It is interesting to note that GeomNet achieves the best accuracy with $(k,k')=(2,3)$, i.e. none of 
$k$ and $k'$ is equal to 1. 
This is opposed to previous works~\cite{GongCVPR09,LiIsSecond17,NguyenHandRecgCVPR19,G2DeNet17}, 
where $n$-variate Gaussians are always identified with 
SPD $(n+1) \times (n+1)$ matrices. To the best of our knowledge, this is the first work that shows 
the benefits of identifying $n$-variate Gaussians with SPD $(n+k) \times (n+k)$ matrices where $k > 1$.     
The results also reveal that the setting of $(k,k')$ has a non-negligible impact on the
accuracy of GeomNet. Namely, the performance gap between 
two settings $(k,k')=(1,1)$ (94.54\%) and $(k,k')=(2,3)$ (96.33\%) is 1.79\%.  
We can also notice that when $k$ is fixed, GeomNet always performs best with $k' > 1$. 
This shows the effectiveness of our parameterization of Riemannian Gaussians in~(\ref{eq:spdstats_embedding}).   

\begin{table}[t]
\begin{center}
  \resizebox{0.85\linewidth}{!}{
  \def\arraystretch{1.2}
  \begin{tabular}{| l | c | c | c |}     
    \hline
    \multirow{2}{*}{{\bf Dataset}} & \multirow{2}{*}{{\bf SBU Interaction}} & \multicolumn{2}{c|}{{\bf NTU RGB+D 60 Dataset}} \\ 
    \cline{3-4}
    & & {\bf X-Subject} & {\bf X-View} \\          
    \hline    
    Without PT & 71.51 & 62.18 & 66.83 \\
    \hline
    PT & {\bf 96.33} & {\bf 93.62} & {\bf 96.32} \\         
    \hline	
  \end{tabular} 
  } 
\end{center}
\caption{\label{tab:exp_ablation_pt} Effectiveness of PT on SBU Interaction and NTU RGB+D 60 datasets.} 
\end{table}

\begin{figure}[t] 
\centering		
\includegraphics[width=0.5\textwidth, trim = 0 60 0 100, clip=true]{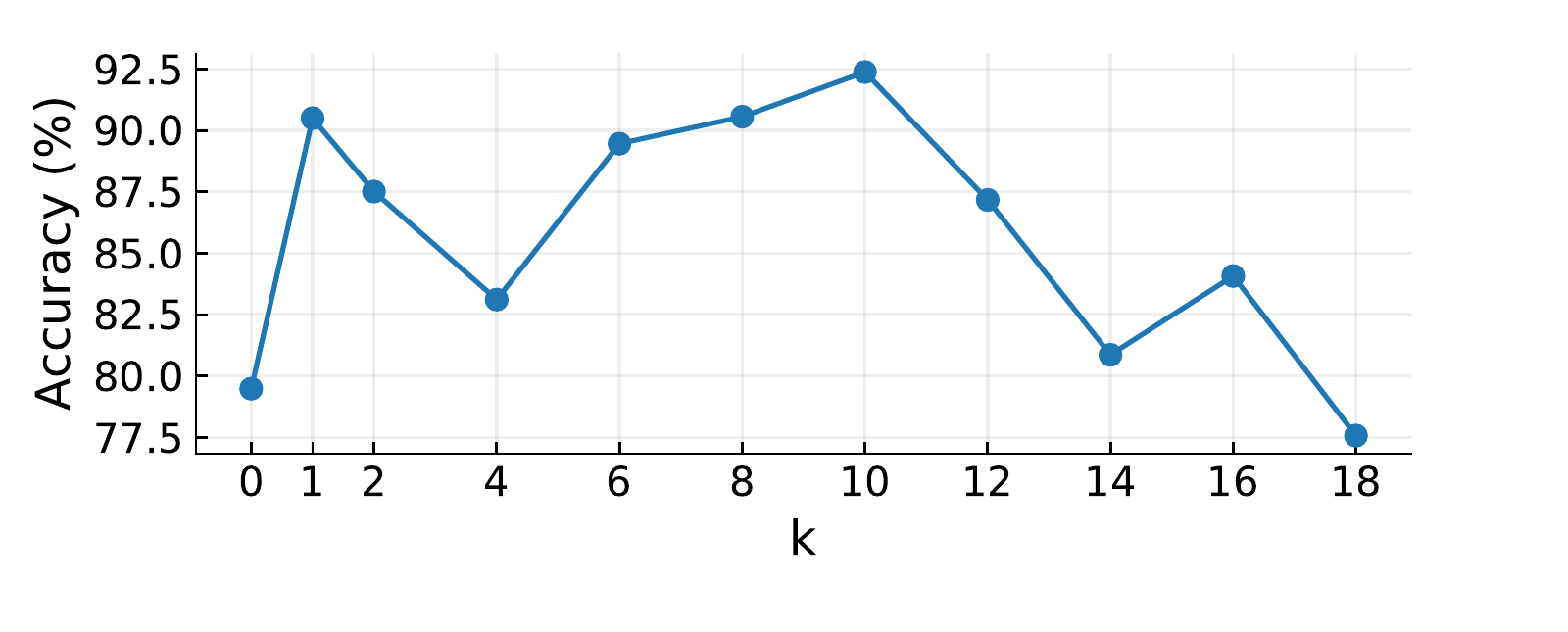}		
\caption{\label{fig_b:accuracy_spdnet} Accuracy of SPDNet on SBU Interaction dataset with different settings of $k$.}
\end{figure}

To investigate the effectiveness of our proposed embedding of Gaussians   
outside of our framework, we used it to improve the state-of-the-art neural network on SPD manifolds 
SPDNet~\cite{HuangGool17}. In~\cite{HuangGool17},  
the authors performed action recognition experiments by representing each sequence by a joint covariance descriptor. 
The covariance descriptor is computed from the second order statistics 
of the 3D coordinates of all body joints in each frame. 
For SBU Interaction dataset, the size of the covariance matrix is $90 \times 90$ (30 body joints in each frame).   
In our experiment, we combined the covariance matrix and the mean vector using the proposed embedding of Gaussians
to represent each sequence. Each sequence is then represented by a SPD $(90+k) \times (90+k)$ matrix. 
We used the code of SPDNet\footnote{\url{https://github.com/zhiwu-huang/SPDNet}}   
published by the authors.  
Fig.~\ref{fig_b:accuracy_spdnet} shows the accuracies of SPDNet\footnote{The results are averaged over 10 runs.} on SBU Interaction dataset with different settings of $k$.   
As can be observed, SPDNet gives the best accuracy with the setting $k=10$. 
The performance gap between two settings $k=1$ (90.5\%) and $k=10$ (92.38\%) is 1.88\%. 
The accuracy of SPDNet when using only the covariance ($k=0$) is 79.48\%, which is significantly worse than
its accuracy with the setting $k=10$. 
The results confirm that our proposed embedding of Gaussians is effective in the framework of SPDNet and that 
it is advantageous over the one of~\cite{Lovric00}. This suggests that our method could also be beneficial to
previous works that rely on
Gaussians to capture local feature distribution, 
e.g.~\cite{GongCVPR09,Li17,LiIsSecond17,Matsukawa16,NguyenHandRecgCVPR19,G2DeNet17}.       

\begin{table}[t]
\begin{center}
  \resizebox{0.9\linewidth}{!}{
  \def\arraystretch{1.2}
  \begin{tabular}{| l | c | c | c |}     
    \hline
    \multirow{2}{*}{{\bf Dataset}} & \multirow{2}{*}{{\bf SBU Interaction}} & \multicolumn{2}{c|}{{\bf NTU RGB+D 60 Dataset}} \\ 
    \cline{3-4}
    & & {\bf X-Subject} & {\bf X-View} \\          
    \hline    
    Without LTML & 94.90 & 92.30 & 95.05 \\
    \hline
    LTML & {\bf 96.33} & {\bf 93.62} & {\bf 96.32} \\         
    \hline	
  \end{tabular} 
  } 
\end{center}
\caption{\label{tab:exp_ablation_tril} Effectiveness of lower triangular matrix learning (LTML) on SBU Interaction and NTU RGB+D 60 datasets.} 
\end{table}

{\bf Parallel transport.} Tab.~\ref{tab:exp_ablation_pt} gives the accuracies of GeomNet without using PT 
on SBU Interaction and NTU RGB+D 60 datasets. The accuracies of GeomNet are also shown for comparison purposes.     
When PT is not used, the covariance in Eq.~(\ref{eq:vectorize_cov}) is computed as: 
\begin{equation}
\mathbf{P}^c = \frac{1}{L-1} \sum_{i=1}^L f_{v}\big(\overrightarrow{\mathbf{P}^m\mathbf{P}_i}\big) f_{v}\big(\overrightarrow{\mathbf{P}^m\mathbf{P}_i}\big)^T. 
\end{equation} 

It can be seen that the use of PT is crucial for obtaining high accuracy. Specifically, on NTU RGB+D 60 dataset, 
computing the covariance without PT 
results in a loss of 31.44\% on X-Subject protocol and a loss of 29.49\% on X-View protocol. 
On SBU Interaction dataset, a significant reduction in accuracy (24.82\%) can also be observed when PT is not used. 
These results highlight the importance of learning the parameters $\mathbf{W}^b_{pt},b=1,2$ in GeomNet.      

{\bf Lower triangular matrix learning.} Tab.~\ref{tab:exp_ablation_tril} gives the accuracies
of GeomNet without using the layer $f_{trilmap}$ on SBU Interaction and NTU RGB+D 60 datasets.  
Again, the accuracies of GeomNet are also shown for comparison purposes.  
We can note that the introduction of the layer $f_{trilmap}$  
brings performance improvement, i.e. 1.43\% on SBU Interaction dataset, and
1.32\% on X-Subject protocol and 1.27\% on X-View protocol on NTU RGB+D 60 dataset.   

\subsection{Results on SBU Interaction Dataset}
\label{sec:exp_sbu_interaction}

Results of GeomNet and state-of-the-art methods on SBU Interaction dataset are given in Tab.~\ref{tab:exp_sbu}. 
For SPDNet, we report its best accuracy using the embedding in~(\ref{eq:embed_final}) with $k=10$. 
We can remark that the accuracies of most of the hand-crafted feature based methods~\cite{JiContrastInteract14,LieGroup14}
are lower than 90\%.  
The state-of-the-art method~\cite{NN15} for skeleton-based action recognition only gives
a modest accuracy of 80.35\%, the second worst accuracy among the competing methods.  
GeomNet achieves the best accuracy of 96.33\%, which is 
16.85\% better than that of SPDNet.     

\begin{table}[t]
\begin{center}
  \resizebox{0.7\linewidth}{!}{
  \def\arraystretch{1.3}
  \begin{tabular}{| l | c |}    
    \hline
    {\bf Method} & {\bf Accuracy}  \\          
    \hline   
    Lie Group~\cite{LieGroup14} & 47.92 \\
    Constrast Mining~\cite{JiContrastInteract14} & 86.90 \\
    Interaction Graph~\cite{LiMultiviewInteract16} & 92.56 \\
    Trust Gate LSTM~\cite{LiuLSTM17} & 93.30 \\ 
    Hierarchical RNN~\cite{NN15} & 80.35 \\
    Deep LSTM+Co-occurence~\cite{ZhuCo-occu16} & 90.41 \\
    SPDNet~\cite{HuangGool17} & 92.38 \\
    \hline                  
    {\bf GeomNet} & {\bf 96.33}  \\         
    \hline	
  \end{tabular}
  } 
\end{center}
\caption{\label{tab:exp_sbu} Recognition accuracy (\%) of GeomNet and state-of-the-art methods on SBU Interaction dataset. 
}
\end{table}

\subsection{Results on NTU RGB+D 60 Dataset}
\label{sec:exp_prediction_ntu60}

Tab.~\ref{tab:exp_ntu60} shows the results of GeomNet and state-of-the-art methods on NTU RGB+D 60 dataset. 
For ST-GCN and AS-GCN, we used the codes\footnote{\url{https://github.com/yysijie/st-gcn}}\textsuperscript{,}\footnote{\url{https://github.com/limaosen0/AS-GCN}} published by the authors.   
For SPDNet, we report its best accuracy using the embedding in~(\ref{eq:embed_final}) with $k=3$. 
We can observe that GeomNet gives the best results on this dataset.  
Since ST-GCN is based on fixed skeleton graphs which might miss implicit joint correlations,
AS-GCN improves it by learning actional links to capture the latent dependencies between joints. 
AS-GCN also extends the skeleton graphs to represent structural links. 
However, AS-GCN does not achieve significant improvements over ST-GCN.          
This indicates that actional and structural links in AS-GCN are still not able 
to cope with complex patterns in 3DTPIR.  
As can be seen, GeomNet outperforms ST-GCN and AS-GCN by large margins. 
We can also note a large performance gap between GeomNet and SPDNet.    
This can probably be explained by the fact that: 
(1) GeomNet aims to learn inter-person joint relationships; 
(2) GeomNet leverages the covariance information on SPD manifolds.  

\begin{table}[t]
\begin{center}
  \resizebox{0.65\linewidth}{!}{
  \def\arraystretch{1.3}
  \begin{tabular}{| l | c | c |}    
    \hline
    {\bf Method} & {\bf X-Subject} & {\bf X-View}  \\          
    \hline    
    ST-LSTM~\cite{LiuTrustGateECCV16} & 83.0 & 87.3 \\    
    ST-GCN~\cite{YanAAAI18} & 86.75 & 91.17 \\
    AS-GCN~\cite{LiActGCN19} & 87.08 & 92.04 \\
    LSTM-IRN~\cite{perez2019interaction} & 90.5 & 93.5 \\   
    SPDNet~\cite{HuangGool17} & 74.85 & 76.07 \\    
    \hline                  
    {\bf GeomNet} & {\bf 93.62}  & {\bf 96.32}  \\         
    \hline	
  \end{tabular}
  } 
\end{center}
\caption{\label{tab:exp_ntu60} Recognition accuracy (\%) of GeomNet and state-of-the-art methods on NTU RGB+D 60 dataset. 
}
\end{table}

\begin{table}[t]
\begin{center}
  \resizebox{0.7\linewidth}{!}{
  \def\arraystretch{1.3}
  \begin{tabular}{| l | c | c |}    
    \hline
    {\bf Method} & {\bf X-Subject} & {\bf X-Setup}  \\          
    \hline    
    ST-LSTM~\cite{LiuTrustGateECCV16} & 63.0 & 66.6 \\   
    ST-GCN~\cite{YanAAAI18} & 78.60 & 79.92 \\
    AS-GCN~\cite{LiActGCN19} & 77.83 & 79.30 \\
    LSTM-IRN~\cite{perez2019interaction} & 77.7 & 79.6 \\   
    ST-GCN-PAM~\cite{YangPairwiseICIP20} & 83.28 & \\
    SPDNet~\cite{HuangGool17} & 60.72 & 62.08 \\       
    \hline                  
    {\bf GeomNet} & {\bf 86.49}  & {\bf 87.58}  \\         
    \hline	
  \end{tabular}
  } 
\end{center}
\caption{\label{tab:exp_ntu120} Recognition accuracy (\%) of GeomNet and state-of-the-art methods on NTU RGB+D 120 dataset. 
}
\end{table} 

\subsection{Results on NTU RGB+D 120 Dataset}
\label{sec:exp_prediction_ntu120}

Results of GeomNet and state-of-the-art methods on NTU RGB+D 120 dataset are given in Tab.~\ref{tab:exp_ntu120}. 
For SPDNet, we report its best accuracy using the embedding in~(\ref{eq:embed_final}) with $k=3$. 
As can be observed, GeomNet performs best on this dataset. 
Note that LSTM-IRN performs significantly worse than GeomNet on this most challenging dataset. 
By adapting the graph structure in ST-GCN to involve connections between two skeletons, 
ST-GCN-PAM achieves significant improvements. 
However, ST-GCN-PAM is still outperformed by GeomNet by 3.21\% on 
X-Subject protocol\footnote{The authors did not report its accuracy on X-Setup protocol.}.  
The results indicate that: 
(1) without any prior knowledge, automatic inference of intra-person and inter-person joint relationships is difficult;     
(2) even with prior knowledge, the state-of-the-art ST-GCN  
performs worse than GeomNet. 
Compared to the results on NTU RGB+D 60 dataset, the performance gap between GeomNet and SPDNet 
is more pronounced on this dataset. 
Notice that our method is based only on the assumption that the joints 
of the arms of two persons and those of their legs are highly correlated during their interaction. 
Therefore, no explicit assumption in pairwise joint connections is required for interaction recognition.   

\section{Conclusion}
\label{sec:conclu}

We have presented GeomNet, a neural network based on embeddings of Gaussians and Riemannian Gaussians for 3DTPIR. 
To improve the accuracy of GeomNet, we have proposed the use of PT and a layer that learns lower triangular matrices 
with positive diagonal entries. 
Finally, we have provided experimental results on three benchmarks showing the effectiveness of GeomNet.  

{\bf Acknowledgments.} We thank the authors of NTU RGB+D datasets for providing access to their datasets. 

{\small
\bibliographystyle{ieee_fullname}
\bibliography{references} 
}

\end{document}